%% file: main.tex
\documentclass{llncs}
\usepackage[hyphens]{url}  
\usepackage[utf8]{inputenc}
\usepackage[dvipsnames,table]{xcolor}
\usepackage{amssymb}
\usepackage{amsmath} 
\usepackage{multirow}
\usepackage{diagbox}
\usepackage{listings}     
\usepackage{graphicx}    
\usepackage{multirow} 

\usepackage{lipsum}
\usepackage{arydshln}
\usepackage{slashbox}

\usepackage{tikz}
\usetikzlibrary{shapes}
\usetikzlibrary{positioning}
\usetikzlibrary{plotmarks}

\lstnewenvironment{code}[1][]
  {\lstset{
   numbers=left,
   xleftmargin=0.6cm,
     numberstyle=\footnotesize,
     escapeinside=@@,
     #1}}
      {}

\newcommand{\itv}[1]{\ensuremath{\mathbf{#1}}}
\newcommand{\mocc}{{\sc globalOCC}}

\newcommand{\restrict}{\textsc{Restrict}}
\newcommand{\default}{\textsc{Baseline}}
\newcommand{\diversify}{\textsc{Diversify}}

\title{Bringing freedom in variable choice when searching counter-examples in floating point programs}
\author{Heytem Zitoun\inst{1} \and Claude Michel\inst{2} \and Laurent Michel\inst{1} \and Michel Rueher\inst{2}}
\institute{University of Connecticut \\\email{ldm@uconn.edu}, 
\email{heytem.zitoun@uconn.edu}
\and
I3S (University Côte d'Azur - CNRS) \\\email{firstname.lastname@i3s.unice.fr}}

\begin{document}
\maketitle

\input{abstract.tex}

\input{intro.tex}

\input{search.tex}

\input{restrict.tex}

\input{diversification.tex}

\input{experiments.tex}

\input{relatedworks.tex}

\input{conclusion.tex}

\bibliographystyle{plain}
\bibliography{cpaior.bib}
\end{document}

%% file: abstract.tex
\begin{abstract}
%
Program verification techniques typically focus on finding counter-examples that violate properties of a program. Constraint programming offers a convenient way to verify programs by modeling their state transformations  and specifying searches that seek counter-examples. Floating-point computations present additional challenges for verification given the semantic subtleties of floating point arithmetic. %
This paper focuses on search strategies for CSPs using floating point numbers constraint systems and dedicated to program verification. It introduces a new search heuristic based on the global number of occurrences that outperforms state-of-the-art strategies. More importantly, it demonstrates that a new technique that only branches on input variables of the verified program
improve performance. It composes with a diversification technique that prevents the selection of the same variable within a fixed horizon  further improving performances and 
reduces disparities between various variable choice heuristics.
The result is a robust methodology that can tailor the search strategy according to the sought properties of the counter example.
\end{abstract}

%% file: intro.tex
\section{Introduction}

Program verification techniques typically focus on finding counter-examples that violate properties of a program. Constraint programming offers a convenient way to verify programs by modeling their state transformations  and specifying searches that seek counter-examples. Floating-point computations present additional challenges for verification given the semantic subtleties of floating point arithmetic. In this context, research strategies play a key role in the effectiveness of the search for counter-examples.

Variable choice strategies and, more generally search heuristics, are a key feature in constraint programming. Those often exploit the semantics of the application or the domain to make smart decisions on what to branch on next. Classic examples of such contributions include \texttt{first-fail}~\cite{Haralick1980}, \texttt{dom/deg}~\cite{Bessiere1996}, \texttt{wdeg}~\cite{Boussemart2004}, \texttt{IBS}~\cite{Refalo2004}, \texttt{ABS}~\cite{Michel2012}, \texttt{Counting}~\cite{Pesant2016} and exploit properties of the CSP such as domain size, static, dynamic or failure-weighted degrees, impact of propagation, activity of variables or number of solutions to name just a few.

The bulk of the efforts in this space was devoted almost exclusively to finite domain search strategies.
%
Reasoning over continuous domains with intervals is well studied too. Systems such as Newton~\cite{VanHentenryck1998}, Numerica~\cite{VanHentenryck1997}, Icos~\cite{LMRDM05} or RealPaver~\cite{Granvilliers2006} are representative examples where the objective is to offer a conservative outer approximation of the reals. The core of the efforts there is predominantly focused on adapting consistency notions to the continuous domain and providing effective filtering operators.
%
In contrast, little attention has been devoted to CSPs using \emph{floating point variables} \cite{MRL01,M02,PMR16}.  Yet, floating point constraint systems are commonly used in the context of program verification 
where one wishes to model the behavior of the system rather than conservatively approximating the reals. 
Floating point domains offer unique properties that could be exploited to enhance the search process. In ~\cite{cp2017}, the authors introduce a collection of dedicated search strategies for program verification that effectively exploit such properties. Those heuristics focused on variable selections as well as branching schemes for splitting domains. The paper concluded that \emph{absorption} and \emph{density} were particularly effective variable selection heuristics and their adoption moved the state of the art forward. 
%
The empirical evaluation in~\cite{cp2017} was based on a small set of benchmarks derived directly from {\tt C} source code of test programs which were manually translated into a system of constraints expressing relationships between states and following Hoare logic~\cite{Hoare1969}. The heuristics included both static and dynamic variants (whose recommendations depend on the current state of the search). 

\emph{This paper revisits search heuristics to solve floating point verification problems and investigates three issues.} 
%
First, it extends the variable selection heuristics to deliver an alternative that significantly improves upon~\cite{cp2017}.
Second, it considers a \emph{simple} static technique called \restrict{} to focus the search on important variables. The technique is particularly potent given its simplicity and  effectiveness. The technique is orthogonal to the variable selection or the splitting strategy. It can, therefore, \emph{be composed with any search heuristic}. Interestingly, \restrict{} does significantly reduce the performance gaps that exist between all the proposed heuristics. 
Finally, the paper amplifies \restrict{} with a simple dynamic diversification technique for the variable selection that is reminiscent of the diversification methods used within meta-heuristics~\cite{Glover1989}. 

This paper includes an evaluation of those techniques over a sizeable and realistic set of benchmarks and demonstrates the value of combining \restrict{} with \emph{diversification}. This combination is extremely promising and deliver a simple yet potent technique that significantly outperforms the results in~\cite{cp2017} while requiring very little implementation efforts. As a matter of facts, combining \restrict{} with \emph{diversification} reduces drastically the gap between the different variable choice strategies and lets the user freely choose the strategy according to other criteria than speed. For instance, the user can select a strategy that promotes solutions in the neighborhood of zero or  absorption according to his needs without regard to efficiency issues.


\paragraph{Contributions} The contributions of this paper include:
\begin{itemize}
    \item \mocc{} a new search heuristic that outperforms state-of-the-art strategies introduced in \cite{cp2017}.
    \item \restrict{} a new technique that only branches on input variables of the program to verify.
    \item \diversify{} a new technique that prevents the selection of the same variable within a fixed horizon (i.e., within a fixed depth of its last selection).
\end{itemize}

    The paper is organized as follows. Section~\ref{sec:heuristics} reviews the properties supporting the variable selection heuristics of~\cite{cp2017} and introduces \mocc{}. Section~\ref{sec:search}
    outlines the generic search procedure. 
    Section~\ref{sec:restrict} introduces \restrict{}. Section~\ref{sec:diversification} presents \diversify{} to further amplify \restrict{}. Section~\ref{sec:expe} presents the empirical evaluation. Section~\ref{sec:rws} discusses related works. 

%% file: search.tex
\section{Heuristic Properties}
\label{sec:heuristics}

Heuristics govern variable selection strategies and order the set of variables according to weights.
Two types of heuristics can be distinguished: dynamic and static (or structural) heuristics.
Width, cardinality, density and absorption belong to the first family as variable weights evolve along with the search tree.
On the other hand, lexicographic,  degree, local occurrences and global occurrences compute variable weights once at the beginning of the search tree.
Note that floating point representation has a significant impact on the behavior of classical heuristics like width or cardinality.
In this section, all property definitions but definition \ref{def:global} are extracted from \cite{cp2017}.

Let the model $M=\langle X,D,C\rangle$ denote a constraint satisfaction problem (CSP) defined over a set of variables $X$, a set of domains $D$ (i.e., floating point intervals) and a set of constraints $C$. Each constraint $c \in C$ is a relation defined over a subset of variables from $X$. Namely, for each $c \in C$, $vars(c) \subseteq X$ represents all the variables occurring in $c$. Likewise, for each variable $x \in X,  cstr(x)$ is the set of constraints in which $x$ appears. Finally, an interval of floating point numbers $\itv{x} = [\underline{\itv{x}}, \overline{\itv{x}}] = \{x \in F, \underline{\itv{x}} \leq x \leq \overline{\itv{x}}\}$, where $F$ is the set of floating point numbers, represents the domain of variable $x$. 

%
%
%
\subsection{Width}
\begin{definition}[Width]
Let $w(\itv{x})$ represent the width of $\itv{x}$, the domain of variable $x$ from the CSP $M=\langle X,D,C\rangle$. Namely,
$
w(\itv{x}) = \overline{\itv{x}} - \underline{\itv{x}}
$ 
in which $\overline{\itv{x}}$ (resp. \underline{\itv{x}}) is the upper (resp. lower) bound of interval $\itv{x}$, i.e., the domain of variable $x$.
\label{def:Width}
\end{definition}

\subsection{Cardinality}
\begin{definition}[Cardinality]
Let $card(\itv{x})$ represent the cardinality (i.e., the number of floating point values) of variable $\itv{x}$. Namely,
\[
card(\itv{x}) = 2^p\times (e_{\overline{\itv{x}}} - e_{\underline{\itv{x}}}) + m_{\overline{\itv{x}}} - m_{\underline{\itv{x}}} + 1
\] 
where $e_{\overline{\itv{x}}}$ (resp.  $e_{\underline{\itv{x}}}$) is the exponent of $\overline{\itv{x}}$ (resp. $\underline{\itv{x}}$), and $m_{\overline{\itv{x}}}$ (resp. $m_{\underline{\itv{x}}}$) is the mantissa of $\overline{\itv{x}}$ (resp. $\underline{\itv{x}}$), while $p$ is the length of the mantissa.
\label{def:cardinality}
\end{definition}

Contrary to what occurs with classical domains like integers, $w(\itv{x})$ and $card(\itv{x})$ capture completely different quantities due to the non-uniform distribution of  floating point numbers. For more details refer to~\cite{cp2017}.

\subsection{Density}
\begin{definition}[Density]
\label{def:Density}
Let  $dens(x)$ represent the density of the domain of variable $x$ from the CSP $M=\langle X,D,C\rangle$. Namely,
\[
dens(x) = \frac{card(\itv{x})}{w(\itv{x})}
\] 
\end{definition}
Observe how the domains of floating-point numbers are not uniformly distributed, e.g., about half of the floats are in $[-1,1]$. Informally, \emph{dens} capture the proximity of floating point in domains. Maximizing \emph{dens} helps capture variables with huge numbers of values in regards to the size of the domain. Intuitively, those variables may have a larger number of values appearing in solutions. 

\subsection{Absorption}
The semantics of floating point arithmetic can surprise developers with behaviors like absorption which is summarized as follows. When two floating point values $a$ and $\epsilon$ are added (or subtracted), if $\epsilon$ is sufficiently small w.r.t. $a$ ($\epsilon$'s magnitude depends on $a$), the result of the operation is $a$ and in this case $\epsilon$ is absorbed by $a$, i.e., $a \pm \epsilon = a$.

\begin{definition}[Absorption]
Let $abs(x)$ represent the maximum number of floats that can be absorbed from the domain of another variable $y$ appearing jointly with $x$ in a constraint of the form $z = x \pm y$. Namely,
\[
abs(x) = \max_{c \in cstr(x) | c \equiv z=x\pm y} absorb(x,y)  
\]
where  $absorb(x,y)$ denote the number of floats of $y$ domain's absorbed by at least a value of $x$ domain's and defined by  

\[
absorb(x,y) = \frac{card([-2^{e_{max} -p -1},2^{e_{max} -p -1}] \cap \itv{y})}{card(\itv{y})}
\]
where p is the size of the mantissa (i.e., 23 for single floating point variable and 53 for double), and $e_{max}$ is the exponent of $\max(|\underline{x}|,|\overline{x}|)$
\end{definition}
The goal of $abs$ is to identify the variable absorbing the most over the constraints. Absorption is a behavior that often lead a program to invalidate its specification. 

\subsection{Lexicographic}
\begin{definition}[lexicographic]
Let $lex(x)$ represent the index of the variable $x$ in the lexicographic order of declaration of the variables. Namely, 
\label{def:lex}
$lex(x_i) = i$
where $i$ represents the $i^{th}$ variable declared. Hence, given $x_i$ and $x_j$ two variables with $i < j$, then $x_i$ has been declared before $x_j$ (i.e, $lex(x_i) < lex(x_j))$.
\end{definition}

\subsection{Degree}
\begin{definition}[Degree]
Let $degree(x)$ denote the degree of a variable $x$ from CSP $\langle X,D,C\rangle$. Namely, $degree(x)$ is the number of constraints referring to $x$:
\[
degree(x) = \sum_{c \in C} \left(x \in vars(c)\right)
\] 
\label{def:degree}
\end{definition}
\subsection{Local occurrence}
\label{sec:local}
Local occurrence focuses 
on each constraint in isolation based on how often a variable appears within the constraint when given in existential form. 

\begin{definition}[Local occurrence]
\label{def:local}
Let $occ_l(x)$ represent the maximum number of occurrences of a variable $x$ in the constraint set $C$ of CSP $\langle X,D,C\rangle$, i.e., 
\[
occ_l(x) = \max_{c \in cstr(X)} \mbox{count}(x,c)
\]
\noindent where $count(x,c)$ is the number of occurrences of $x \in c$. 
\end{definition}
For instance, if $c$ is a polynomial, $count(x,c)$ is the sum of the degrees of $x$ in the terms of $c$.
Intuitively, the expression of a constraint $c\in C$ can include multiple references to the same variable $x$. The property $occ_l(x)$ imputes to variable $x$ the largest number of occurrences among all the constraints in the CSP. 

\subsubsection{Example}
\label{sec:local:ex}

$M=\langle X=\{x,y,z,w\},D=\{\itv{x}=\itv{y}=\itv{z}=\itv{w}=[-10,10]\},C\rangle$ with 
\[
C = \{z = (x + y) \times x, z = y + 1, w = y - 1 \}
\]
Let us detail the computation of $occ_l(x)$ and $occ_l(y)$. Since, $cstr(x)=\{z = (x + y) \times x\}$ and $x$ appears twice in this constraint, $occ_l(x)=2$. Likewise, $cstr(y)=C$ and $y$ appear once in each constraint. Thus, $occ_l = [x \mapsto 2,y \mapsto 1,z \mapsto 1,w \mapsto 1]$ and the local occurrence heuristic will favor variable $x$.

\subsubsection{Limitation} 
 Fundamentally, the property  prefers a variable with more impact within a particular constraint over a variable with impact on the whole problem. For instance, with the set $C$ of Example~\ref{sec:local:ex}, the variable $x$ will be chosen (it occurs twice in the first constraint), even if it belongs to a single constraint. Yet, $y$ occurs three times in the system but only once in each constraint and fixing $y$ leads to a solution by propagation. The situation can be even worse if one ignores the first constraint. In that case, all variables appear at most once in the remaining two constraints and local occurrence will not be able to discriminate.

\subsection{Global occurrence}
\label{sec:global}
The limitation of $occ_l$ can be removed with a  revision to the definition that takes into account the occurrences of a variable in the entire CSP.  

\begin{definition}[Global occurrence]
\label{def:global}
Let  $occ_g(x)$ represent the maximum number of occurrences of a variable $x$ in the CSP $M=\langle X,D,C\rangle$. Namely,
\[
occ_g(x) = \sum_{c \in cstr(x)} \mbox{count}(x,c)
\]
\noindent where $count(x,c)$ is borrowed from definition~\ref{def:local}.
\end{definition}
Intuitively, fixing the variable appearing the most in the whole program might have the most impact on the domain size of all other variables\footnote{The heuristic is \emph{static} as its definition does not depend on the domains of variables. Future work could focus on
blending $occ_g$ with dynamic properties.}.

\subsubsection{Example}
Consider again Example~\ref{sec:local:ex}. 
The $occ_g$ property is therefore $[x \mapsto 2,y\mapsto 3,z \mapsto 2,w \mapsto 1]$. 
%
Global occurrence overcomes the \emph{Local Impact} limitation of local occurrence (i.e., the propensity to favor variables with high local impact over variables affecting constraints throughout the model). 

\section{Search Heuristic}
\label{sec:search}

The adoption of a property $f$ to drive the variable selection heuristic and 
solve a CSP $M=\langle X,D,C\rangle$ amounts to choose the variable $x_k \in X$ that maximizes $f(x)$. In other words, given
$f^* = \max_{x \in X} f(x)$, let the subset of candidate variables be
\[
cand(f^*) = \{ y \in X : f(y) = f^* \}
\]
and the branching variable $x_k$ be drawn from $cand(f^*)$. Whenever
$|cand(f^*)|~>~1$, the generic strategy breaks ties with a lexicographic ordering that leverages the static order of variable declarations in the model.

All the variable selection strategies in~\cite{cp2017} as well the new \mocc{} (which implements definition \ref{def:global} of section \ref{sec:global}) result from the instantiation of the generic search strategy above with one of the properties given in Section~\ref{sec:heuristics}.



%% file: restrict.tex
\section{Restriction}
\label{sec:restrict}
Exploiting the semantics of variables, domains and constraints can be quite effective to tackle hard CSPs. \emph{Yet, exploiting the semantics of the application domain itself} can also prove extremely potent. 
This section discusses a simple technique that takes advantage of the source program being verified to identify and focus the search on a subset of critical variables for a dramatic impact on performance. 

A CSP $M=\langle X,D,C\rangle$ is derived through an automatic process from a {\tt C} program $P$. This \emph{derivation} process weaves the state changes of the program through constraints connecting the input variables of the function $P$  to the output (returned value) of the program. Clearly, when the program $P$ has inputs $\mathcal{I}$ with $\mathcal{I} \subseteq X$ and outputs $O \in X$, looking for a counter-example that violates the post-condition of $P$ amounts to execute a search procedure based on a strategy $\mathcal{S}$ that branches on the variables in $X$. 

\restrict{} is a \emph{simple} departure from this classic approach. Fundamentally, \restrict{} solves $M$ by \emph{exclusively} branching on the input variables of $P$, i.e., it only branches on $\mathcal{I}$ and not the entire set $X$. This does not jeopardize the correctness nor the completeness of the approach. Indeed, once the input variables are fixed, all the auxiliary variables get fixed by virtue of propagation alone. 

It is worth noting that the implementation is straightforward and completely orthogonal to the search strategy (variable selection and branching heuristics) making it trivial to integrate restrict with any search strategy and evaluate its impact on all the alternatives that are available. 

This approach seems quite natural but it is not common in program verification where most tools are based upon SMT solvers. The point is there is no  notion of input variables in the SMT format even if, in the case of problems on floats, the input variables can be identified.

\paragraph{Formalization}
The \emph{derivation} of $M=\langle X,D,C\rangle$ is a compilation of the sequence of instructions appearing in program $P$. First, consider a single instruction.

\begin{example}[State Derivation]
Consider a 3-operand instruction of the form $x = y \diamond z$ in which $\diamond$ is a binary operator and the set of variables representing the state of the program at the start of line 2 includes $\{x_k,y_k,z_k\}$, i.e., 
\begin{code}
... 
x = y @$\diamond$@ z;
...
\end{code}
The assignment on line 2 introduces a fresh variable $x_{k+1}$ and a constraint 
$x_{k+1} = y_k \diamond z_k$
to represent the state transformation in which $x_{k+1}$ is the variable holding the \emph{current} state of the symbolic variable $x$ after the instruction.
\end{example}
The derivation can be generalized to all instructions and applied on entire blocks of code to yield a CSP $\langle T,D,C\rangle$ in which $T$ contains all the temporary variables modeling the state evolution (and their domains) and $C$ is the set of constraints connecting all the states.

\begin{definition}[Verification CSP]
A {\tt C} function
$P(\vec{\mathcal{I}}) \rightarrow O$ is subject to a pre-condition  $Pre(\vec{\mathcal{I}})$ and must satisfy a post-condition $Post(O)$.  In Hoare's terms \[
\{Pre(\vec{\mathcal{I}})\} O = P(\vec{\mathcal{I}}); \{Post(O)\}
\]
captures the expectations on $P$. Given the CSP $\langle T,D,C\rangle$ derived from the body of $P$, the CSP 
\[
F = \langle 
\mathcal{I} \cup T,  
D,
C \cup \{Pre(\vec{\mathcal{I}})\} \cup \{\neg Post(O)\}
\rangle
\]
has a solution $\sigma$ if and only if there exist an assignment of values to the variables in $\mathcal{I}$ that delivers an output $O$ that violates the post-condition, namely, such a $\sigma$ is a counter-example. Similarly, if $F$ has no solution, no such counter-example exists. 
\end{definition}
Traditionally, one would solve the CSP $F$ in search of a solution (a counter-example) and this is exactly what~\cite{cp2017} does. Namely, it applies a branching strategy $\mathcal{S}$ to all the variables of $F$
\[
\textsc{Classic}(F) = solve(F,search(vars(F),\mathcal{S}))
\]
in which {\tt solve} refers to the resolution of the CSP using a {\tt search} procedure based on the branching strategy $\mathcal{S}$ applied to the variables in $F$, i.e., in $vars(F)$. 
\restrict{} is a \emph{simple} departure from this classic approach. It considers the resolution of $F$ by \textit{exclusively} branching on variables from $\mathcal{I}$. More formally, 

\begin{definition}[\restrict{}]
Let $F$ be the verification CSP for a program $P$ whose inputs is $\mathcal{I}$ (recall that $\mathcal{I} \subseteq vars(F)$). $\restrict{}(F)$ only considers variables from $\mathcal{I}$ in its search procedure, namely
\[
\restrict{}(F) = solve(F, search(\mathcal{I},\mathcal{S}))
\]
\end{definition}

%% file: diversification.tex
\section{Diversification}
\label{sec:diversification}

Diversification techniques come from meta-heuristics. A classic example is Tabu search~\cite{Glover1989} which insists on preventing the repetition of local moves that \emph{undo recent changes} and induce cyclic behaviors where the same computation state is visited repeatedly. Tabu is a diversification mechanism that insists on shifting the focus to moves involving other variables, thereby driving the search in a different part of the search space. 
In that context, diversification is often considered in an alternating pattern alongside intensification to sample the search space more extensively and sometimes focus on areas of interest. 

The idea has rarely 
been considered in the context of a complete search tree\footnote{To our knowledge, only \cite{JUSSIEN2000} proposed to use a tabu to handle nogoods in the context of a path-repair algorithm.}. Note that, there is no risk to ``cycle'' and revisit the same configuration in a tree search and completeness guarantees that the whole space will be inspected, explicitly or implicitly. 
Yet, when considering floating-point variables, the idea has a natural appeal. Consider a variable with a large domain and a simple bisection heuristic for branching. If the variable selection heuristic finds the variable appealing (e.g., because of its occurrence counter), the search may become obstinate and repeatedly branch on the variable. Since the domain density is not uniform\footnote{Note that while the interval $[0, 1]$ contains about a quarter of the floats, $[10^5, 10^5+1]$ contains only $128$ simple floats.}, such obstinacy may very well lead the search to a great depth before switching to another variable. It may, therefore, be desirable to forbid the selection of this variable in the near future if it was already branched on recently. 

More formally, each time a variable $x$ is selected for branching at some tree node $n$ (at depth $depth(n)$), the search records a prohibition depth for $x$ equal to 
\[
last(x) = depth(n) + u
\]
where $u$ sets a desired depth of search for which the variable is prohibited.

Intuitively, the search is prohibited to consider $x$ again until the depth exceeds $last(x)$.
When the search considers a  tree node $n$ (at depth $depth(n)$), the selection heuristic, rather than inspecting any (free) variable in $X$, may instead choose to consider
\[
Y = \mathcal{S}(\{ y \in X | \neg bound(y) \wedge last(y) \leq depth(n) \}).
\]
where $bound(y)$ is true if variable $y$ is bounded.

Clearly, as the value of the parameter $u$ increases, variables stay in ``purgatory'' increasingly longer after being selected. With $u=0$, this strategy simply reduces to normal branching and one must always have $u \leq |X|$ to operate normally\footnote{In practice, if $u$ exceeds the limit, it is set to the limit.
}. Note that, as the search backtracks, the prohibition depths must be restored. Finally, observe how the heuristic $\mathcal{S}$ is applied to the non-prohibited variables to yield a branching recommendation.





%% file: experiments.tex
\section{Experiments}
\label{sec:expe}
This section evaluates \mocc,  \restrict{} and \diversify{} on a significant set of benchmarks and discusses their impacts and interplay.
\vspace{-2mm}
\paragraph{Benchmarks}
The following experiments consider 151 benchmarks with 84 benchmarks with solutions (SAT), and 67 without solution (UNSAT).  Those extend significantly~\cite{cp2017} and originate from standard program verification's sources. More precisely, benchmarks include {\tt C} code from section 6.3 of~\cite{Damouche2016} (i.e., main contribution of FP\_Bench), and benchmarks from SMTLIB~\cite{smtlib} where the {\tt C} code is available.
Specifically, they include:
\begin{itemize}
  \item All the benchmarks described in~\cite{cp2017}.
  \item LeadLag, PID, Odometrie, Runge Kutta (second and fourth order) and Trapeze programs are all coming from~\cite{Damouche2016}. Two versions of these programs are available: a classic 
    and an \emph{optimized} one with a reduced amount of numerical error. The benchmarks compare the difference between the classic and optimized programs and  try to find
    input values where the square of the difference is bigger or equal to a given delta. Different benchmarks are obtained by unfolding the inner loop from 1 to 200 times.
  \item Other benchmarks are from the SMTLIB~\cite{smtlib}, namely  add*, div*, 
mul*, e1\_*, e2\_*, newton*, sine*, square\_*. The benchmarks from the SMTLIB are restricted to the one with available {\tt C} source.
\end{itemize}

%
%

Overall, these benchmarks have up to $2813$ variables and up to $2813$ constraints.
With respect to variables, $40$ benchmarks have between $10..100$ variables, 
$26$  benchmarks have  $100..1000$ variables,
and $3$ of them have in excess of a $1000$ variables.
The cardinality of the restricted set of variables reaches up to $51$ variables.
As a matter of fact, \restrict{} reduces the set of variables taken into account by the search by more than $90\%$ for $43$ benchmarks, by $50\%$ up to $90\%$ for $77$ benchmarks and, by less than $50\%$ for $31$ benchmarks. Note that $2$ benchmarks do not benefit from \restrict{}.

All the experiments were made on a Macbook Pro with 2.9GHz 6-core Intel Core i9 and 32GB with a timeout of one minute. Note that all experiments have been done using the split 5 way strategy described in~\cite{cp2017}.

\vspace{-2mm}
\paragraph{Raw Results}
Tables~\ref{tab:full} reports, for each strategy from~\cite{cp2017} (in light gray) as well as for the new strategy \mocc{}, the number of timeouts as well as the total runtime for instances that are satisfiable (SAT), unsatisfiable (UNSAT) or both (ALL). The timeout is set to 60 seconds. The four \emph{sections} of the table report the results for:
\begin{description}
    \item[\default{}] the implementation in~\cite{cp2017}. Namely, the results in the section \default{} for each strategy are from~\cite{cp2017} evaluated on the extended benchmark collection proposed here. 
    \item[\restrict{}] the \default{} augmented with the static restriction technique only.
    \item[\diversify{}] the \default{} augmented with the dynamic diversification technique only (with $u = 2$). 
    \item[\restrict{}+\diversify{}] the \default{} augmented with both the static restriction and the dynamic diversification techniques.
\end{description}
Column \mocc{} reports the result for the new strategy. The last column of the table conveys the standard deviation of the time ($\sigma(t_s)$) across the population of all strategies (i.e., 8 strategies) as well as the mean time ($\mu(t_s)$).
\begin{table*}[t]
    \centering
    {\scriptsize
    \begin{tabular}{|c|c|c|c|c|c|c|c|c|c||c||c|}
    \cline{4-11}
        \multicolumn{3}{c}{} & \multicolumn{8}{|c|}{Strat.}  \\
    \hline
        \multirow{2}{*}{Feature} & \multirow{2}{*}{Type} & \multirow{2}{*}{} & max & max  & Local & max & max & max & lex. & Global & \multicolumn{1}{l|}{\hspace*{0.0cm}$\sigma(t_s)$} \\ 
        & &  & Dens & Card & Occ & Deg & Width & Abs & & Occ & \multicolumn{1}{r|}{$\mu(t_s)$\hspace*{0.0cm}}\\
        \hline
        \multirow{6}{*}{\default{}} & \multirow{2}{*}{SAT} & To & \cellcolor{Gray!15}18  &  \cellcolor{Gray!15}37 & \cellcolor{Gray!15}11 & \cellcolor{Gray!15}8 & \cellcolor{Gray!15}34 & \cellcolor{Gray!15}19 & \cellcolor{Gray!15}6 & \textbf{1} & \multicolumn{1}{l|}{745.06} \\ 
        && $t_s$ & \cellcolor{Gray!15}1084.89  & \cellcolor{Gray!15}2312.91 & \cellcolor{Gray!15}707.14 & \cellcolor{Gray!15}509.50 & \cellcolor{Gray!15}2097.27 & \cellcolor{Gray!15}1144.64 & \cellcolor{Gray!15}407.20 & \textbf{92.67} & \multicolumn{1}{r|}{\hspace*{0.25cm}1044.53}\\

        &\multirow{2}{*}{UNSAT}&To & \cellcolor{Gray!15}35  & \cellcolor{Gray!15}32 & \cellcolor{Gray!15}9 & \cellcolor{Gray!15}\textbf{6} & \cellcolor{Gray!15}29 & \cellcolor{Gray!15}13  & \cellcolor{Gray!15}12 & \textbf{6} & \multicolumn{1}{l|}{647.21} \\ 
        &&$t_s$ & \cellcolor{Gray!15}2110.83  & \cellcolor{Gray!15}1962.95 & \cellcolor{Gray!15}661.03 & 
\cellcolor{Gray!15}\textbf{469.04} & \cellcolor{Gray!15}1795.53 & \cellcolor{Gray!15}883.98 & \cellcolor{Gray!15}830.34 &  \textbf{461.12} & \multicolumn{1}{r|}{1146.85} \\

        &\multirow{2}{*}{ALL}&To & \cellcolor{Gray!15}53  & \cellcolor{Gray!15}69 & \cellcolor{Gray!15}20 &\cellcolor{Gray!15}14 & \cellcolor{Gray!15}63 & \cellcolor{Gray!15}32  & \cellcolor{Gray!15}18 & \textbf{7} &  \multicolumn{1}{l|}{1323.60} \\ 
        &&$t_s$ & \cellcolor{Gray!15}3195.72  & \cellcolor{Gray!15}4275.86 & \cellcolor{Gray!15}1368.17 & \cellcolor{Gray!15}978.54 & \cellcolor{Gray!15}3892.8 & \cellcolor{Gray!15}2028.62 & \cellcolor{Gray!15}1237.54  & \textbf{553.79} & \multicolumn{1}{r|}{2191.38}\\
        \hline
        \multirow{6}{*}{\restrict} & \multirow{2}{*}{SAT} & To & 1  &  \textbf{0} & 2 & \textbf{0} & \textbf{0} & 1 & 1 & \textbf{0} &
        \multicolumn{1}{l|}{41.45} \\ 
        
        && $t_s$ & 94.59  &  \textbf{35.13} & 154.43 & \textbf{35.1}3 & \textbf{34.81} & 94.16  & 94.50 & \textbf{35.38} & \multicolumn{1}{r|}{72.27}\\

        &\multirow{2}{*}{UNSAT}&To & 6  & \textbf{5} & 6 & 6 & \textbf{5} & 6 & 6 & 6 & \multicolumn{1}{l|}{51.92} \\ 
        
        &&$t_s$ & 474.22  & \textbf{352.14} & 474.34 & 463.26 & \textbf{351.74} & 478.51 & 476.46 & 459.19 & \multicolumn{1}{r|}{441.23}\\

        &\multirow{2}{*}{ALL}&To & 7  &  \textbf{5} & 8 & 6 & \textbf{5} & 7  & 7 & 6 & \multicolumn{1}{l|}{83.50} \\ 
        
        &&$t_s$ & 568.80 & \textbf{387.28} & 628.77 & 498.39 & \textbf{386.55}  & 572.67 & 570.96 & 494.57 & \multicolumn{1}{r|}{513.50}\\
        \hline
        \multirow{6}{*}{\diversify{}} & \multirow{2}{*}{SAT} & To & 15   &  30 & 9 & 10 & 31 & 14 & \textbf{1} & 2& \multicolumn{1}{l|}{625.13} \\ 
        
        && $t_s$ &  905.99  & 1840.13 & 580.07 & 607.65 & 1871.62 & 856.5  & \textbf{133.64} & 173.33 & \multicolumn{1}{r|}{871.12} \\

        &\multirow{2}{*}{UNSAT}&To & 32  &  32 & \textbf{5} & \textbf{5} & 25 & 10 &  \textbf{5} & \textbf{5}&   \multicolumn{1}{l|}{683.99} \\ 
        
        &&$t_s$ & 1932.15  & 1924.36 & \textbf{373.67} & \textbf{354.88} & 1557.98 & 665.47 & \textbf{370.49} & \textbf{352.86} & \multicolumn{1}{r|}{941.48}\\

        &\multirow{2}{*}{ALL}&To & 47  & 62  & 14 & 15 & 56 & 24 & \textbf{6} & 7 & \multicolumn{1}{l|}{1244.39} \\ 
        
        &&$t_s$ & 2838.15 & 3764.49 & 953.74 & 962.53 & 3429.59 & 1521.96 & \textbf{504.13} &526.19&  \multicolumn{1}{r|}{1812.57}\\
        \hline
         & \multirow{2}{*}{SAT} & To & 1  &  0 & 0 & 0 & 0 & 0 & 0 & 0  & \multicolumn{1}{l|}{19.86} \\ 
        \restrict{}&& $t_s$ & 94.46  & 34.29 & 34.54 & 34.35 & 34.5 & 34.21  & 34.43 & 34.46  & \multicolumn{1}{r|}{41.91}\\

        \multirow{2}{*}{$+$}&\multirow{2}{*}{UNSAT}&To & 5  & 5  & 5 & 5 & 5 & 5 & 5 & 5 & \multicolumn{1}{l|}{1.39} \\ 
        
        &&$t_s$ & 351.47  & 353.25 & 354.29 & 355.12 & 352.82 & 354.38 & 350.82 & 352.56 & \multicolumn{1}{r|}{353.09}\\

        \diversify{}&\multirow{2}{*}{ALL}&To & 6  & 5 & 5 & 5 & 5 & 5 & 5 & 5 &
        \multicolumn{1}{l|}{19.29} \\ 
        &&$t_s$ & 445.93  & 387.55 & 388.83 & 389.47 & 387.32 & 388.59  & 385.26 & 387.02 & \multicolumn{1}{r|}{395}\\
        \hline
    \end{tabular}
    }
    \caption{Full comparison}
    \label{tab:full}
\end{table*}

\begin{figure}
    \centering
\includegraphics[scale=0.32]{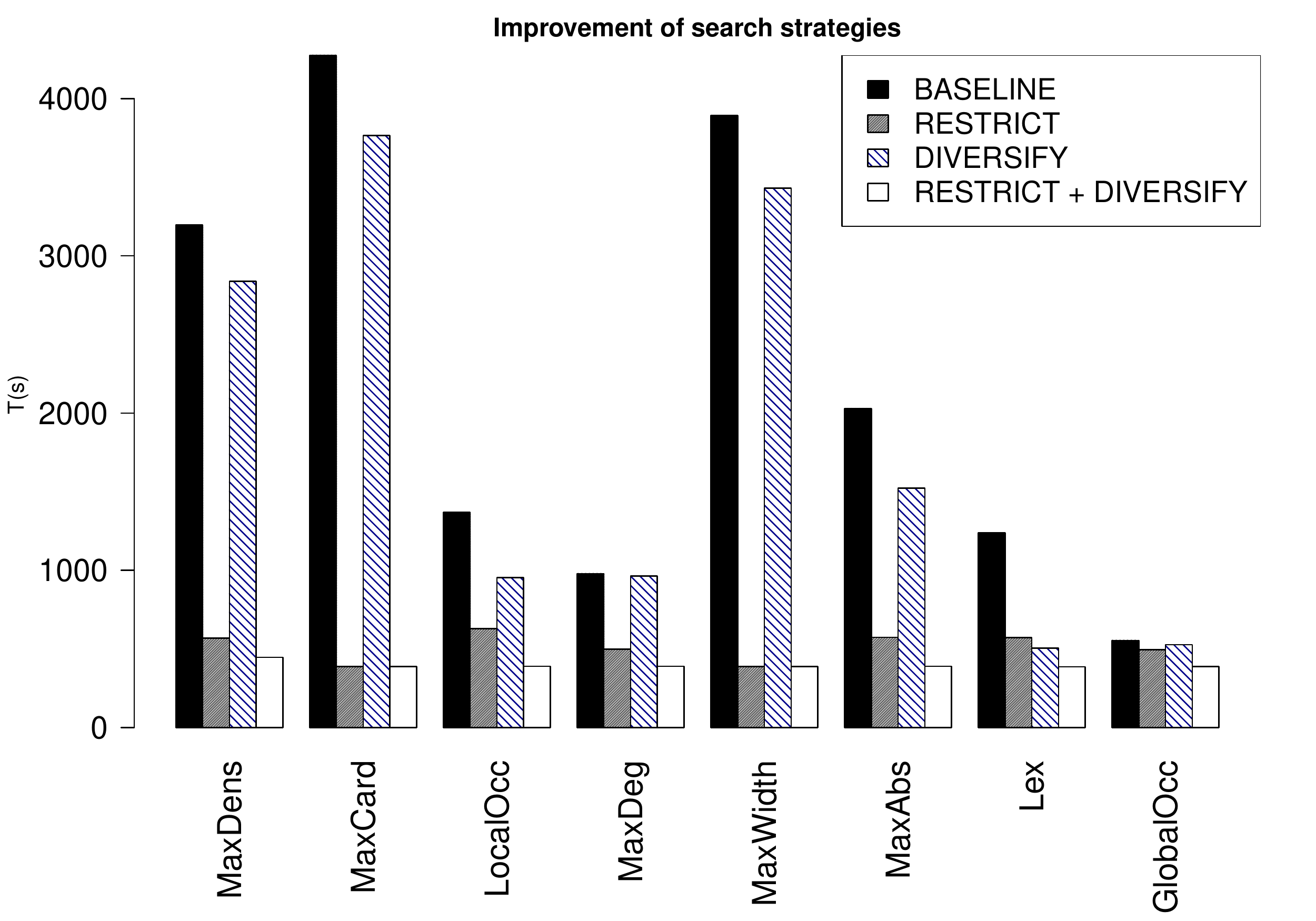}
    \caption{Comparison of features over all benchmarks}
    \label{fig:all}
\end{figure}

\vspace{-2mm}
\paragraph{Impact \& Discussion (\mocc{})}
The \default{} section of table~\ref{tab:full} highlights the benefit of \mocc{} over other strategies when used with the default configuration.
It improves the solving of the satisfiable benchmarks, i.e., the benchmarks which are the most sensitive to the search strategies, by a
factor from $4.39$ up to $24.95$ depending on the variable choice heuristic while keeping ahead of unsatisfiable benchmark solving.

\mocc{} better anticipates the impact of a given variable on the CSP. Contrary to other structural heuristics like degree or local occurrences that consider the local role of a variable, it weighs a variable according to its global effect on the CSP. 
As structural heuristics performs better than dynamic heuristics, \mocc{} dominates all other heuristics.

\vspace{-2mm}
\paragraph{Impact \& Discussion (\restrict{})}
\restrict{} has a dramatic impact on performance. It reduces the number of instances that suffer timeouts and it cuts back on the runtime of the others significantly. While $\mathcal{I}$, the subset of input variables, is contained in $X$, the full set of variables,  $\mathcal{I}$ can remain sizeable. Indeed, functions whose inputs are \emph{arrays} of floating points yield large numbers of input variables as there is one input variable for each entry of the array. Naturally,
if $|\mathcal{I}|=1$, the problem derived from $P$ is uninteresting from a search prospective since there is no choice to make for which variable to branch on. 

\restrict{} delivers a second surprise as the performance gaps that exist between various heuristics when using a search on $X$ shrink dramatically  as one adopts \restrict{}. The gains are so significant that they erase almost entirely the advantage that the best heuristics have when considering the full set of variables $X$. 
Table~\ref{tab:full} provides the necessary evidence. The number of timeouts from \default{} drops dramatically when \restrict{} is used. In aggregate (across all heuristics), \default{} delivers 244 timeouts while \restrict{} produces only 51. The drop in running times is equally impressive and improves \emph{for all heuristics} bringing \default{}'s range of $553.79..4275.86$ seconds 
down to $386.55..572.67$ seconds. 
The ``gap'' between heuristics captured by the standard deviation across the heuristic population drops accordingly from $1323.6$ seconds down to $83.5$ seconds.

A natural question that arises is \emph{Why is \restrict{} so effective?} 
Two situations are worth considering: satisfiable and unsatisfiable instances. 
Recall that satisfiable instances produce a counter-example that violates the program specification. Yet, all the ``internal'' variables (temporaries) are functionally dependent on the input variables. As soon as the inputs are fixed, all the internal state is fixed. Focusing on the inputs is going to quickly reduce the size of the search space. While branching on outputs might sound attractive, the lack of \emph{back-propagation}\footnote{The filtering process used here relies on a local consistency \cite{BGM06}.} may prevent any form of informed guidance. 

Unsatisfiable instances may, on the contrary, look credible until one realizes that the values leading to the counter-examples are not allowable w.r.t. the pre-conditions of the program. As a result, branching on ``internal'' variables is liable to cause the exploration of a very large tree in which candidate solutions are ultimately rejected when the pre-condition on the inputs are finally considered. 

\vspace{-2mm}
\paragraph{Impact \& Discussion (\diversify{})}
\diversify{} is quite simple to implement requiring minimal changes to the search. It is effectively orthogonal to the selection strategy $\mathcal{S}$. As Figure~\ref{fig:all} demonstrates, \diversify{} improves \default{} as well as \restrict{}.
Yet, \diversify{} serves
as an amplification to \restrict{}. Indeed, \restrict{}'s job is to prohibit the selection of temporaries for branching while \diversify{} prohibits excessive focus on the same variables during the search. Combined, they narrow the performance gaps between  different heuristics, 
reduce the number of timeouts even further, and lower the gaps between search heuristics significantly. For instance, {\tt maxCard} drops from $3764.49$ seconds to $387.55$ seconds. 
Even \mocc{} benefits from this combination.
%

Consider the code example shown in Figure~\ref{fig:code}. Lines 2-3 of  function \textbf{f} specify pre-conditions, and lines 7-11 corresponds to post-condition checks. Recall that the derivation builds a CSP $M$ where post-conditions are negated. A correct solution of this program is a pair of values $\langle x,y \rangle$, both in $[-3.0,3.0]$ that leads the execution either to the else block (lines 9-10) or to the if block (lines 7-8) without triggering the assertion. A counter example would lead the code execution to the if block and trigger the assertion. Few counter examples exist for this program.

With the  \texttt{maxDens} strategy, neither \default{} nor \restrict{} find a solution  in a reasonable time (i.e., before the timeout). Both \diversify{} and \diversify{} +\restrict{} find a solution quickly. The latter explore 10 times fewer nodes than \diversify{} alone. On this program, \diversify{} refutes sub-trees substantially faster. More precisely, \diversify{} helps to prune branches that corresponds to $x \leq 0$ in 44 nodes. Whereas \restrict{} still struggle on the same part after almost 2 million nodes. Even the addition of $x \leq 0$  to the model is not enough to get \restrict{} or \default{} to compete.

\begin{figure}[tb]
\begin{lstlisting}[
    language=c,
    basicstyle=\scriptsize,
    frame=lines, 
    numbers=left,
  breaklines=true 
%   postbreak=\mbox{\textcolor{red}{$\hookrightarrow$}\space}
]
float f(float x, float y) {
  assume(-3.0f <= x && x <= 3.0f);
  assume(-3.0f <= y && y <= 3.0f);
  float xy = (x * y),xx = (x * x),yy = (y * y);
  float sum = xx + yy;
  if(3.0f * xy <= (x*xx)+(y*yy) && sum*(yy + (x*(x+1.0f))) <= xy*4.0f) {
    assert(0.100000001f > sum);
    ...
  } else 
    ...
}
\end{lstlisting}
\vspace{-5mm}
\caption{Example of code where \diversify{} improves the search of a counter-example.}
\label{fig:code}
\vspace{-2mm}
\end{figure}

Figure~\ref{tree:comp} illustrates changes in the search process when \diversify{} is added. Here, both searches first branch on variable $x$ with $D(x)~=~[L..U]$ and $Mid=\frac{L+U}{2}$ and carry out a 5-way split. The trees rooted at $L$, $Mid$ and $U$  impose simple branching constraints such as $x=L$, $x=Mid$ and $x=U$ and all are easily proved as finitely failed. However, when imposing $L^+~\leq~x~\leq~Mid^-$, the two search heuristics exhibit different behaviors. \restrict{} (shown in black) favors repeatedly branching on $x$ (because of a high value density) and \emph{dives} in a large sub-tree before reaching a point where another variable $y$ has a better density and gets branched on. Shortly after, the propagation discovers the infeasibility and backtracks. This process leads to a large refutation proof that consumes the entire runtime. In contrast, \restrict{}+\diversify{} is \emph{forced} to skip over the most attractive variable ($x$) and instead favor a second best variable $y$. This alternation between $x$ and $y$ may repeat, but leads to sub-domains for $x$ and $y$ of comparable sizes for which the inconsistency is discovered at a much shallower depth. Because the height $h_1$ of the red tree is much smaller than $h_2$, the refutation completes quickly and the search finally explores the last sub-tree where the solution lies. 
Benchmarks were \diversify{} helps often exhibit this class of behavior which suggest that further work is warranted to automatically determine a suitable value for the ``tabu'' parameter. 

\begin{figure}[tb]
\begin{center}
\input{diversify-drawing.tex}
\end{center}
\vspace{-5mm}
\caption{\restrict{} vs.  \restrict{}+\diversify{} search trees on benchmark \texttt{f23}.}
\label{tree:comp}
\vspace{-4mm}
\end{figure}

\vspace{-2mm}
\paragraph{Hardness}
 
 The benchmarks considered for those experiments are realistic. For instance, with \default{}, the average of total fails for each strategy exceeds 39 millions. The combination of \restrict{} with \diversify{} reduce the size of the explored search tree, yet, the search remains challenging with an average of total fails for all strategies exceeding 7.8 millions.
 
 
\vspace{-2mm}
\paragraph{Summary}
Two key results emerge from the empirical evaluation. 
First, the new heuristic \mocc{}  outperforms the state of the art from~\cite{cp2017}. As Table~\ref{tab:full} demonstrates, its performance exceeds all other heuristics by a significant margin, being at least 56\% faster and up to 772\% faster with the lowest number of timeouts. 
Second, both \restrict{} and \diversify{} have a significant impact on performance. In particular, \restrict{} speeds up all the heuristics. Both are simple to implement, carry no overhead and are orthogonal to the search heuristic. Nonetheless, the techniques are  effective and Figure~\ref{fig:all} leaves no doubt about their value. Indeed, even the worst strategy becomes competitive when \restrict{} and \diversify{} are leveraged. 


%% file: diversify-drawing.tex
\begin{tikzpicture}[level 1/.style={sibling distance=30mm},scale=0.75,every node/.style={scale=0.75}]
\tikzstyle{every path}=[very thick]

\edef\sizetape{0.7cm}

\begin{scope}
\node (z){x}
        child {node  (N0) {$L$}}
        child {node  (N1) {$Mid$}}
        child {node  (N2) {$U$}}
        child {node (N3) {$[L^+,Mid^-]$}}
        child {node (N4) {$[Mid^+,U^-]$}};
\end{scope}

\foreach \t in {0,1,2}
{
\begin{scope}
[shift={(-6+\t*3,-3cm)}]
\node[regular polygon,regular polygon sides=3, draw=black, thick, minimum size=1cm] (fsbox\t)
		{};
\foreach \x in {0,...,2}
\draw node at (-0.35+\x/3,-0.25) {\footnotesize \textcolor{red}{X}};
\end{scope}
}

\begin{scope}
[shift={(1.5cm,-5cm)}]
\node[regular polygon,regular polygon sides=3, draw=black, thick, minimum size=5cm] (fsbox3)
		{};
\foreach \x in {0,...,16}
\draw node at (-2+\x/4,-1.2) {\textcolor{red}{X}};

\draw[<->] (-3,-1.2) --  (-3,2.3);
\draw node at (-4.3,1) {
\begin{tabular}{c}
$h_2$ \\
NO \diversify{}
\end{tabular}};

\draw (0,2.5) .. controls  (.2,2) .. (-.1,1.5) .. controls (-1,1) .. (-.5,0) .. controls (-1,-.5) .. (-.5,-1.1);
\draw node at (-0.5,0.45) {$x$} ;
\draw node at (-0.6,.9) {$x$} ;
\draw node at (0.2,1.5) {$x$} ;
\draw node at (-0.1,1.9) {$x$} ;
\draw node at (-.1,1.2) {$x$} ;
\draw node at (-0.2,0) {$y$} ;
\draw node at (-.5,-.45) {$x$};
\draw node at (-.5,-.9) {$y$};
\end{scope}

\begin{scope}
[shift={(5cm,-4cm)}]
\node[regular polygon,regular polygon sides=3, draw=red, thick, minimum size=1.5cm] (fsbox3-2)
		{};
\foreach \x in {0,...,3}
\draw node at (-0.5+\x/3,-0.4) {\textcolor{red}{X}};

\draw[<->,draw=red] (-1.5,-0.2) --  (-1.5,1);
\draw node at (-1.8,0.5) {
\begin{tabular}{c}
\textcolor{red}{$h_1$} \\
\textcolor{red}{with \diversify{}}
\end{tabular}};

\draw[-,draw=red] (-0.2,-0.4) .. controls (.1,0) .. (0,.8);
\draw node at (0.2,0.8) {\textcolor{red}{$y$}} ;
\draw node at (0.2,0.5) {\textcolor{red}{$x$}} ;
\draw node at (0.2,0.1) {\textcolor{red}{$y$}} ;
\draw node at (0.1,-0.2) {\textcolor{red}{$x$}} ;
\end{scope}

\draw node at(5,-5)  {$h_1 << h_2$};

\begin{scope}
[shift={(6,-3cm)}]
\node[regular polygon,regular polygon sides=3, draw=black, thick, minimum size=1cm] (fsbox4)
		{};
\draw node at (0,-0.3) {\footnotesize  \textcolor{blue}{O}};
\draw (0,.5) .. controls (-.3,0) and (.6,-.1) .. (0,-0.3) ;
\end{scope}

\foreach \id in {0,...,4}
\path[-,draw] (fsbox\id.north)  -- node[right] 
 			{} 
			(N\id.south);

\path[-,draw=red] (fsbox3-2.north)  -- node[right] 
 			{} 
			(N3.south);
\end{tikzpicture}

%% file: relatedworks.tex
\vspace{-2mm}
\section{Related works}
\label{sec:rws}

\paragraph{Variable selection strategies} have received much attention in finite and in continuous CSPs.
Most of the strategies mentioned in the introduction are dedicated to finite domains where small sized domains are the norm. Yet, in program verification and floating point problems in particular, one must handle huge domains.
For continuous domains, the most common variable selection strategy is the Largest First (LF) strategy \cite{Kearfott:1987:TGB:29380.29862,CR97,lg2012}.
It consists in selecting the variable with the domain of maximal width. More sophisticated approaches \cite{lg2012,ArayaN18} rely on the  Maximal Smear (MS) strategy introduced by \cite{Kearfott:1996:RGS,Kearfott:1990:AIP}. Informally, the Maximal smear select the variable the projection of which has the strongest slope.
While continuous domains are often large, the most interesting techniques rely on mathematical properties that do not hold in floating point arithmetic.

In contrast, search heuristics exposed in this paper exploit either of the floating point peculiarities (like density or absorption) or program verification features (like the \restrict{} set).

\vspace{-2mm}
\paragraph{Floating point constraint solvers} have been developed to address verification problems of programs with floating point computations \cite{MRL01,BGM06}.
Available floating point solvers can be divided into two categories: constraint solvers and SMT solvers.
Our approach belongs to the first family of solvers as Colibri~\cite{colibri} does.
Both solvers rely mainly on interval computations adapted to floating point arithmetic to preserve the set of solutions \cite{M02,BGM06}.
A coarse evaluation\footnote{On the sizeable subset of identical benchmarks available for both Colibri and our own implementation.} 
of both solvers showed that Colibri is, on average, twenty percent slower than our solver  in the \default{}  configuration.

SMT solvers like MathSat~\cite{mathsat5}, Z3~\cite{z3} and CVC4~\cite{cvc4}  are now able to handle floating point problems thanks to the QF\_FP theory\footnote{See \url{http://smtlib.cs.uiowa.edu}.}. They are organized around a DPLL \cite{dpll62} procedure which send the floating point subproblem to a dedicated solver.
  SMT solvers are often built on top of SAT solvers and handle floating point arithmetic by means of bit blasting.
  As a result, Colibri outperforms these solvers on QF\_FP benchmarks\footnote{See \url{https://smt-comp.github.io/2019/results/qf-fp-single-query} for SMTCOMP 2019 (note that par4 is a portfolio of solvers) or \url{http://smtcomp.sourceforge.net/2018/results-QF_FP.shtml?v=1531410683} for SMTCOMP 2018.}.
  SMT solvers depend on the SMT language~\cite{smtlib} to describe the problem. Unfortunately, this language does not permit to distinguish input and auxiliary variables.

%% file: conclusion.tex
\vspace{-2mm}
\section{Conclusion}
\label{sec:ccl}
This paper consideres floating point CSPs and offers two contributions. First, it introduces a new search heuristic that considers a \emph{global} occurrence property that substantially improves the state of the art results in this domain. 
Second, it defines two techniques that affect the search strategies yet remain orthogonal to the variable selection or the domain splitting and are therefore composable with any heuristic. 
\restrict{} focuses the search on the subset of variables that appear in the input of the program under verification. It delivers a significant performance boost both in runtime and a dramatic reduction in the number of instances that timeout regardless of the heuristic being employed. 
\diversify{} is inspired by meta-heuristics and prevents the selection of a variable (for branching) if it has been recently branched on. The diversification amplifies the gains of \restrict{} with whom it composes easily. The combination of both \restrict{} and \diversify{} reduces the differences of efficiency between search strategies to almost nothing and thus, brings to the user the freedom of choosing search strategies according to other criteria.
%
Experiments on a significant set of benchmarks are very promising.